\documentclass[11pt,letterpaper]{article}
\usepackage[left=1in,right=1in,top=1in,bottom=1in,letterpaper]{geometry}
\usepackage{amsmath,amsfonts}
\usepackage{bm}      % for \bm macro
\usepackage{todonotes}
\usepackage{float}
\usepackage{textcomp}
\usepackage{xcolor}
\usepackage{graphicx}
\usepackage{booktabs}
\usepackage{multirow}
\usepackage{threeparttable}
\usepackage{epstopdf}
\usepackage{enumerate}
\usepackage{algorithmic}
\usepackage[ruled,longend]{algorithm2e}
\usepackage{hyperref}

\allowdisplaybreaks % from the amsmath package, allows multiline equations to break across pages (delete if not wanted)
\newcommand{\bn}{\bm{n}}
\newcommand{\bg}{\bm{g}}
\newcommand{\bmf}{\bm{f}}
\newcommand{\R}{\mathbb{R}}
\newcommand{\M}{\bm{M}}

\newcommand{\K}{\bm{K}}

\newcommand{\vo}{\bm{0}}

\newcommand{\norm}[1]{\lVert#1\rVert}

\DeclareMathOperator{\diag}{diag}
\begin{document}

\title{Symmetry-Reduced Physics-Informed Learning of Tensegrity Dynamics}

\author{Jing Qin
\thanks{Department of Mathematics, University of Kentucky, Lexington, KY}
\and
Muhao Chen
\thanks{Department of Mechanical and Aerospace Engineering, University of Houston, TX}
}

\date{}

\maketitle
\begin{abstract}
Tensegrity structures possess intrinsic geometric symmetries that govern their dynamic behavior. However, most existing physics-informed neural network (PINN) approaches for tensegrity dynamics do not explicitly exploit these symmetries, leading to high computational complexity and unstable optimization. In this work, we propose a symmetry-reduced physics-informed neural network (SymPINN) framework that embeds group-theory-based symmetry directly into both the solution expression and the neural network architecture to predict tensegrity dynamics. By decomposing nodes into symmetry orbits and representing free nodal coordinates using a symmetry basis, the proposed method constructs a reduced coordinate representation that preserves  geometric symmetry of the structure. The full coordinates are then recovered via symmetry transformations of the reduced solution learned by the network, ensuring that the predicted configurations automatically satisfy the symmetry constraints. In this framework, equivariance is enforced through orbit-based coordinate generation, symmetry-consistent message passing, and physics residual constraints. In addition, SymPINN improves training effectiveness by encoding initial conditions as hard constraints, incorporating Fourier feature encoding to enhance the representation of dynamic motions, and employing a two-stage optimization strategy. Extensive numerical experiments on symmetric T-bars and lander structures demonstrate significantly improved prediction accuracy and computational efficiency compared to standard physics-informed models, indicating the great potential of symmetry-aware learning for structure-preserving modeling of tensegrity dynamics.
\end{abstract}
%% abstract minimum 200 words


\section{Introduction}

% connelly2022frameworks
Tensegrity structures are spatial frameworks composed of isolated compression members embedded within a continuous network of tension elements \cite{skelton2009tensegrity}.
In a tensegrity system, compression members do not directly contact one another but are stabilized by a surrounding tension network, resulting in lightweight yet mechanically efficient structures \cite{fraddosio2024fast}.
Due to their high strength-to-weight ratio, deployability, and structural adaptability, tensegrity systems have attracted considerable attention in aerospace engineering, robotics, and architectural design \cite{charandabi2025tensegrity,ding2025high,ma2025equilibrium}.
The mechanical behavior of tensegrity structures is governed by highly nonlinear dynamics, subject to geometric constraints, prestress states, and coupled interactions among structural members \cite{ma2020design}. These nonlinear characteristics make the accurate simulation of tensegrity dynamics computationally challenging. Traditional approaches, e.g., finite element methods, often require significant computational effort, particularly when repeated simulations are required for structural design, control, or real-time prediction \cite{ma2022tensegrity}.
 % and dynamic relaxation techniques

In recent years, {physics-informed neural networks} (PINNs) have emerged as a promising alternative framework for solving differential equations using deep learning. Different from traditional discretization-based methods, PINNs incorporate governing physical laws directly into the training process of neural networks by embedding the residuals of the governing equations into the loss function \cite{raissi2019physics,karniadakis2021physics}. This approach enables neural networks to learn solutions that are consistent with both the observed data and the underlying physical laws. By incorporating the governing equations directly into the training process, PINNs can effectively combine data-driven modeling with physics-based constraints. As a result, they have shown strong potential for analyzing and designing tensegrity structures. Recent studies have successfully applied PINNs to problems such as tensegrity form-finding \cite{wang2025two} and the design and control of tensegrity systems \cite{zhu2025physics}.

% , and prediction of tensegrity dynamics \cite{xu2025predicting}.

Applying PINNs to tensegrity dynamics provides a flexible framework for learning structural responses while enforcing physical laws \cite{xu2025predicting}. However, several challenges arise when applying standard PINNs to solve tensegrity dynamical systems. First, the dimensionality of the variables in the governing equation increases with the number of nodes, resulting in high-dimensional learning problems that significantly increase the computational burden of training. Second, the nonlinear and strongly coupled interactions among structural members can lead to stiff dynamical behavior, which may slow down or destabilize the optimization process. Third, conventional PINN formulations typically ignore intrinsic structural symmetries, resulting in redundant representations that fail to exploit the system's underlying geometric structure. Finally, the vanilla PINN framework has limited capability to accurately capture long-time or periodic dynamics when only a small number of training data points are available. To address these challenges, it is useful to exploit intrinsic structural properties of the system.

% tibert2003review
One distinctive feature of many tensegrity systems is the presence of {geometric symmetry}. Regular tensegrity configurations, such as prisms, towers, and spherical tensegrity structures, often exhibit cyclic or dihedral symmetries due to repeated arrangements of nodes and structural members. These symmetries can be described mathematically through group actions involving permutations of nodes together with spatial transformations. Exploiting symmetry has been recognized as an effective strategy for simplifying structural analysis. In particular, symmetry can reduce the dimensionality of the configuration space and enable decomposition of structural equations into symmetry-adapted subspaces. Such ideas have been extensively studied in the theory of symmetric frameworks and applied to tensegrity systems for equilibrium analysis, form-finding, and stability analysis \cite{connelly1998mathematics,connelly2022frameworks}.
In practical applications, symmetry constraints have been shown to reduce the number of design variables in form-finding methods such as the force density method or energy-based equilibrium formulations \cite{raj2006using,chen2018improved}. Recently, symmetry-based reinforcement learning has been proposed to improve tensegrity locomotion performance \cite{surovik2021adaptive}. %These techniques enable analytical characterization of symmetric equilibrium states in prismatic tensegrity structures and facilitate the study of stiffness and vibration properties.
Despite these advances, most existing work focuses on analytical modeling or structural optimization, while the integration of symmetry principles into modern learning-based approaches for tensegrity dynamics remains relatively limited.

Motivated by these observations, we propose an efficient symmetry-reduced physics-informed neural network (SymPINN) framework for modeling tensegrity dynamics that explicitly incorporates structural geometric symmetry into the learning process. By identifying symmetry transformations via node permutations, together with spatial operations such as rotations and reflections, we characterize the admissible configurations of a tensegrity structure as a symmetry-invariant subspace and construct a corresponding basis to represent the structural coordinates using a reduced set of variables. A physics-informed neural network is then trained in this reduced coordinate space to approximate the time evolution of these variables while enforcing the governing physical laws through physics-based loss functions. This formulation offers three main advantages: it reduces the dimensionality of the learning problem, improving computational efficiency and training stability; it ensures that predicted solutions automatically satisfy the symmetry constraints; and it improves interpretability by revealing the fundamental motion patterns associated with the symmetry group of the tensegrity system.

The rest of the paper is organized as follows. In Section~\ref{sec:sym}, we describe symmetric tensegrity structures and their mathematical representations. In Section~\ref{sec:method}, we reformulate the governing equation in the symmetry subspace, and propose a physics-informed neural network approach by embedding the tensegrity symmetry. Numerical experiments on various data sets are presented in Section~\ref{sec:exp}, together with a brief discussion of method configurations. Finally, we conclude the paper in Section~\ref{sec:con}.

%%%%%%%%%%%%%%%%%%%%%%%%%%%%%%%%%%%%%%%%%%%%%%%%%%%%%%%%%%%
\section{Symmetry of Tensegrity Structures}\label{sec:sym}

In many real-world applications, tensegrity structures have symmetry to some extent, which can be utilized to further simplify the modeling and reduce computational complexities. Specifically, symmetry in tensegrity refers to a set of geometric transformations (motions) that map a tensegrity configuration onto itself without changing its equilibrium state. It can be described by a finite symmetry group acting on the set of nodes and edges of the structure. In \cite{connelly1998mathematics}, symmetry is used to classify prismatic tensegrities and simplify the stiffness matrix and equilibrium equations. In \cite{connelly2022frameworks}, symmetry in tensegrities is analyzed using group representation theories.

The primary symmetry operations encountered in tensegrity structures arise from point groups in Euclidean space. The collection of all operations that leave a tensegrity configuration unchanged forms a symmetry group, which is typically a cyclic or dihedral point group. In this work, we focus on these two common types of symmetry.

\subsection{Cyclic Symmetry}\label{sec:c_sym}

A tensegrity structure is said to possess {cyclic symmetry of order $n$} if its configuration is invariant under rotation by an angle $2\pi/n$ about a specified axis, where $n$ is a positive integer. Given the nodes of the tensegrity structure as $\{\bm{q}_1,\ldots,\bm{q}_m\}$, we define the nodal coordinate matrix as
\[
\bm{Q}=[\bm{q}_1,\ldots,\bm{q}_m]\in\mathbb{R}^{3\times m}.
\]
Let
$
\bm R_z(\alpha)\in\mathbb{R}^{3\times3}
$
denote the rotation matrix corresponding to a rotation by angle $\alpha$ about the $z$-axis. The structure is said to have {$\mathcal{C}_n$ symmetry} if the rotation
$
\bm R := \bm R_z\!\left({2\pi}/{n}\right)
$
maps the set of nodes onto itself. Since the nodes may be relabeled after the rotation, this invariance can be expressed as
$
\bm R\bm{Q}=\bm{Q}\bm{P}
$,
where $\bm{P}\in\mathbb{R}^{m\times m}$ is a permutation matrix representing the induced relabeling of nodes.
Equivalently, the nodal configuration is invariant under the cyclic group
\[
\mathcal{C}_n=\left\{\bm R_z\!\left(\frac{2k\pi}{n}\right):k=0,1,\ldots,n-1\right\},
\]
which consists of all rotations generated by repeated application of $\bm R_z(2\pi/n)$. Cyclic symmetry commonly occurs in tensegrity structures composed of identical modules arranged around a central axis, such as prismatic towers and ring-like configurations.

\subsection{Dihedral Symmetry}

A tensegrity structure is said to possess {dihedral symmetry of order $2n$} if it is invariant under the action of the dihedral group
\[
\mathcal{D}_n=\langle \bm R,\bm S\rangle
:=
\{\bm I,\bm R,\bm R^2,\ldots,\bm R^{n-1},\bm S,\bm S\bm R,\bm S\bm R^2,\ldots,\bm S\bm R^{n-1}\},
\]
which consists of $n$ rotations and $n$ reflections. Here $\bm I$ is an identity matrix with the same size as $\bm R$, $\bm R=\bm R_z(2\pi/n)$ generates the rotational subgroup $\mathcal{C}_n$, and $\bm S\in\mathbb{R}^{3\times3}$ denotes a reflection with respect to a plane containing the symmetry axis, for example, the $xz$-plane. The tensegrity structure with nodes as columns of the matrix $\bm Q$ is said to have {$\mathcal{D}_n$ symmetry} if for every group element $\bm B\in \mathcal{D}_n$, it holds
$
\bm B\bm{Q}=\bm{Q}\bm{P}_{\bm B}
$,
for some permutation matrix $\bm{P}_B$. This condition implies that the structure is invariant under both rotations by multiples of $2\pi/n$ and reflections across planes passing through the symmetry axis. Many prismatic tensegrity structures formed by two parallel $n$-node rings connected by vertical or inclined members naturally exhibit $\mathcal{D}_n$ symmetry.

% \subsection{Linear Symmetry Constraint}

% For computational purposes, it is convenient to express the symmetry condition in vector form. Let
% \[
% \bm{q}=\operatorname{vec}(\mathbf{Q})
% \in\mathbb{R}^{3m}
% \]
% denote the stacked vector of nodal coordinates. Using the property
% \[
% \operatorname{vec}(R\mathbf{Q})=(I_m\otimes R)\bm{q},
% \qquad
% \operatorname{vec}(\mathbf{Q}\mathbf{P})=(\mathbf{P}^\top\otimes I_3)\bm{q},
% \]
% the cyclic symmetry condition $R\mathbf{Q}=\mathbf{Q}\mathbf{P}$ can be rewritten as
% \[
% (I_m\otimes R - \mathbf{P}^\top\otimes I_3)\bm{q}=0.
% \]

% Let
% \[
% S := I_m\otimes R - \mathbf{P}^\top\otimes I_3.
% \]
% Then the set of all configurations satisfying the symmetry constraint forms the nullspace, denoted by $\ker(S)$.
% If $U\in\mathbb{R}^{3m\times r}$ is a matrix whose columns form an orthonormal basis of $\ker(S)$, any symmetric configuration can be expressed as
% \[
% \bm{q}=U\bm{z},
% \]
% where $\bm{z}\in\mathbb{R}^r$ represents the reduced coordinates. This representation provides the foundation for the symmetry-reduced dynamic model and the associated physics-informed neural network formulation developed in the following sections.

\section{Symmetry-Reduced Physics-Informed Learning}\label{sec:method}
Suppose a tensegrity structure consists of $n_e$ elements, including both bars and strings, and $n_n$ nodes in total. Among these nodes, $n_a$ are free nodes and $n_b$ are constrained nodes, with $n_n = n_a + n_b$. The all nodal coordinates at time $t$ are stacked into a column vector $\bm{n}(t) \in \mathbb{R}^{3n_n}$, which collects the three-dimensional coordinates of all nodes. Furthermore, by denoting the free nodal coordinates by $\bm{n}_a\in\R^{3n_a}$ and the constrained nodal coordinates by $\bm{n}_b\in\R^{3n_b}$, we have
\begin{equation}
\bm{n}=\bm{E}_a\bm{n}_a+
\bm{E}_b\bm{n}_b,
\end{equation}
where $\bm{E}_a\in\R^{3n_n\times 3n_a}$ and $\bm{E}_b\in\R^{3n_n\times 3n_b}$ are the respective indicator matrices for the free and constrained nodes. In addition, we make the following assumptions on the tensegrity structure: (a) the structure is symmetric with respect to the nodal coordinates as well as the string and bar connectivity; (b) the constrained nodes are arranged symmetrically so that the constraints are invariant under the symmetry transformation; (c) gravity acts in the vertical direction and therefore preserves the symmetry of the structure; (d) the external forces are symmetric; (e) the initial conditions are symmetric; and (f) the force densities and material properties are identical for members belonging to the same symmetry orbit.

% In addition, we make the following assumptions on the tensegrity structure:
% \begin{enumerate}[(a)]\setlength{\itemsep}{0pt}
% \item the tensegrity structure is symmetric with respect to the nodal coordinates as well as the string and bar connectivity;
% \item the constrained nodes are arranged symmetrically so that the constraints are invariant under the symmetry transformation;
% \item gravity acts in the vertical direction and therefore preserves the symmetry of the structure;
% \item the external forces are symmetric;
% \item the initial conditions are symmetric;
% \item the force densities and material properties are identical for members belonging to the same symmetry orbit.
% \end{enumerate}

\subsection{Symmetry-Reduced Dynamic Equation}
Consider a nonlinear tensegrity dynamic equation of the following form \cite{ma2022tensegrity}:
\begin{equation}\label{eqn:dyn_vec}    \bm{E}_a^T(\bm{M}\ddot{\bn}+\bm{D}\dot{\bn}+\bm{K}(\bn,t)\bn-\bmf_{ex}(t)+\bg)=\vo,
\end{equation}
where $\vo$ denotes the zero vector, $\bm{M}$, $\bm{D}$, and $\bm{K}$ are the respective mass, damping, and stiffness matrices, $\bmf_{ex}$ stands for the external force, and $\bg$ corresponds to the gravity. Specifically, $\M=\M_s\otimes \bm I_3$, $\K=\K_s\otimes \bm I_3$ and $\bg$ are defined as follows:
\begin{align}
&\bm{M}_s =\frac {1} {6}\Big(|\bm{C}|^T\diag({\bm{m}})|\bm{C}|+ \lfloor|\bm{C}|^T\diag({\bm{m}})|\bm{C}|\rfloor\Big),\label{eq:M}\\
&\bm{K}_s =(\bm{C}^T\diag(\bm{x}(\bm{n},t))\bm{C}),\label{eq:K}\\
&\bm{g} = \frac{g}{2}(|\bm{C}|^T\bm{m})\otimes[\,0~~0~~1\,]^T,\label{eq:g}
\end{align}
where $\bm{C}\in\R^{n_e\times n_n}$ is the connectivity matrix, $\bm{m}\in\R^{n_e}$ is the mass vector, $\bm{I}_3$ is the 3-by-3 identity matrix, $\bm{x}\in \R^{n_e}$ is the force density vector depending on $\bn$ and $t$, $|\cdot|$ is an operator to get the absolute value componentwise, $\diag$ returns a diagonal matrix whose diagonal entries are from the vector, $\lfloor \cdot \rfloor$ zeroes out the off-diagonal entries, and  $\otimes$ is the Kronecker product.
Because the element force density $\bm{x}$ depends on the current configuration and time, $\bm{K}$ is generally nonlinear in $(\bm{n},t)$. The initial conditions are defined as
\begin{equation}\label{eqn:ic}
    \bm{n}(0)=\bm{\phi},\quad \dot{\bm{n}}(0)=\bm{\psi},
\end{equation}
where $\bm{\phi}$ and $\bm{\psi}$ represent the initial coordinates and velocity, respectively.
Then the problem boils down to solving the second-order dynamical system for $\bm{n}_a$. By denoting
\[
\begin{aligned}
    \bm M_a&=\bm E_a^T M\bm E_a, ~
    \bm D_a =\bm E_a^T D\bm E_a, ~
    \bm K_a=\bm E_a^T K\bm E_a,\\
    \bm w_a&=\bm E_a^T (-\bm M\bm E_b\ddot{\bm n}_b-\bm D\bm E_b\dot{\bm n}_b-\bm K\bm E_b\bm n_b+\bm f_{ex}-\bm g),
\end{aligned}
\]
we get a reduced system for the free nodes
\begin{equation}\label{eqn:dyn_na}
\bm M_a \ddot{\bm n}_a+\bm D_a\dot{\bm{n}}_a+\bm K_a\bm n_a -\bm w_a=\bm{0}.
\end{equation}
The initial conditions are reduced to
\begin{equation}
\begin{aligned}
    \bm{n}_a(0)&=\bm \phi_a:=\bm E_a^T(\bm \phi-\bm E_b \bm{n}_b(0)),\\
    \dot{\bm{n}}_a(0)&=\bm \psi_a:=\bm E_a^T(\bm \psi-\bm E_b\dot{\bm n}_b(0)).
\end{aligned}
\end{equation}

Next, we reshape the nodal coordinates as a matrix and then derive the symmetry condition and symmetry basis.
Let $\bm{N}\in\mathbb{R}^{n_a\times 3}$ denote the free nodal coordinate matrix whose vectorized form satisfies $\bm{n}=\operatorname{vec}(\bm{N}^T)$, where $\operatorname{vec}(\cdot)$ converts a matrix to a vector by columnwise stacking. Due to the symmetry assumption for the tensegrity structure, there exists a permutation matrix $\bm P\in\R^{n_n \times n_n}$ and an orthogonal matrix $\bm R\in\R^{3\times 3}$ corresponding to a rotation or a composition of rotation with reflection such that
\begin{equation}\label{eqn:check_sym}
(\bm{N}-\bm{1}\bm{c}^T)\bm R=\bm P(\bm{N}-\bm{1}\bm{c}^T),
\end{equation}
where the center of symmetry $\bm{c}\in\R^{3}$ is the mean coordinates of all the nodes in $\bm{N}$ and $\bm{1}\in\R^{n_n}$ is a all-one vector. For the simplicity of discussion, we assume $\bm{c}=\bm{0}$ without loss of generality, since the symmetry can always be applied to coordinates relative to the center $\bm{c}$.
In practice, this condition can be used to check the dynamic nodal symmetry.

Under the symmetry assumptions, the node set of a tensegrity structure can be partitioned into multiple {orbits}, where each orbit is a subset of nodes that are mapped to one another under the action of the symmetry group. Let $n_r$ denote the number of representative nodes, i.e., one node selected from each orbit, with $n_r \le n_a$. For example, in a $C_3$ prism tensegrity consisting of six nodes arranged in two rings, the symmetry partitions the nodes into two orbits, i.e., top ring and bottom ring, and thus the number of representative nodes $n_r=2$.

To obtain a reduced representation of the solution, we introduce a symmetry basis, defined as the subspace of coordinates that remain invariant under the symmetry transformation described in \eqref{eqn:check_sym}. By taking the transpose of the symmetry relation $\bm N \bm R = \bm P\bm N$ and moving the right term to the left, we get its equivalent form
\[
\bm R^T\bm N^T \bm I_{n_a}-\bm I_3\bm N^T \bm P^T=\bm O.
\]
Here $\bm O$ is the $3\times n_a$ zero matrix.
Using the property of Kronecker products
\begin{equation}\label{eqn:kron_prop}
    (\bm A\otimes \bm B)\operatorname{vec}(\bm X)=\operatorname{vec}(\bm B\bm X\bm A^T),
\end{equation}
we have
\[\begin{aligned}
\operatorname{vec}(\bm R^T\bm N^T\bm I_{n_a})&=\bm I_{n_a}\otimes \bm R^T\operatorname{vec}(\bm N^T),\\
\operatorname{vec}(\bm I_3\bm N^T \bm P^T)&=\bm P\otimes \bm I_3\operatorname{vec}(\bm N^T).
\end{aligned}
\]
Since $\operatorname{vec}(\bm N^T)=\bm n_a$, we obtain
\[
(\bm I_{n_a}\otimes \bm R^T - \bm P\otimes \bm I_3)\bm{n}=\bm{0}.
\]
This shows that the symmetry-invariant coordinate vectors belong to the nullspace of the matrix
\begin{equation}\label{eqn:S}
\bm S:=\bm I_{n_n}\otimes \bm R^T-\bm P\otimes \bm I_3.
\end{equation}
Therefore, the symmetry basis is any basis spanning the nullspace of $\bm S$. The dimension of this space equals the number of independent symmetric degrees of freedom, i.e., the number of symmetry orbits. Since both $\bm P$ and $\bm R$ are orthogonal matrices, their eigenvalues lie on a unit circle in the complex space $\mathbb{C}$. By taking the respective eigendecompositions of $\bm P$ and $\bm R^T$, we find their common eigenvalues, denoted by $\lambda_i$. Assume $\bm{v}_i$'s are the orthonormal eigenvectors of $\bm P$ and $\bm{w}_i$'s are the orthonormal eigenvectors of $\bm R^T$ such that
\[
\bm P\bm{v}_i=\lambda_i\bm{v}_i \quad\mbox{and}\quad \bm R^T\bm{w}_i=\lambda_i\bm{w}_i.
\]
Then the vectors $\bm{u}_i=\bm{v}_i\otimes \bm{w}_i$ form a symmetry basis.
In the special case where the symmetry involves only a permutation matrix $\bm P$, the symmetry basis can be obtained from the orthonormal eigenvectors of $\bm P$ corresponding to the eigenvalue one.

Let $\bm{U}\in\mathbb{R}^{3n_a\times n_r}$ denote the matrix whose columns form the symmetry basis. Using this basis, the nodal coordinate vector can be expressed as
\begin{equation}\label{eqn:rec_n}
\bm{n}_a(t)=\bm{U}\bm{z}(t) \quad\mbox{and}\quad \bm{z}(t)\in\R^{n_r}.
\end{equation}
Since $\bm U^T\bm U=I_{n_r}$, we have $\bm U^T\bm n_a(t) = \bm z(t)$. Substituting this representation into the governing dynamics for the free nodes \eqref{eqn:dyn_na} followed by multiplying $\bm U^T$ on the left yields the reduced system
\begin{equation}\label{eqn:dyn_sym}
    \bm{M}_r \ddot{\bm{z}}+\bm{D}_r \dot{\bm{z}}+\bm{K}_r \bm{z}-\bm{w}_r=\vo.
\end{equation}
Here $\bm{M}_r=\bm{U}^T\bm{M}_a\bm{U}$, $\bm{D}_r=\bm{U}^T\bm{D}_a\bm{U}$, $\bm{K}_r=\bm{U}^T\bm{K}_a\bm{U}$ and $\bm{w}_r=\bm U^T\bm{w}_a$. The reduced initial conditions are
\begin{equation}\label{eqn:dyn_sym_ic}
{\bm z}(0)=\bm U^T \bm \phi_a \quad\mbox{and}\quad \dot{\bm z}(0)=\bm U^T \bm \psi_a.
\end{equation}
With this formulation, the neural network is trained to learn the reduced coordinates $\bm{z}(t)$. Since the dimension of the reduced system satisfies $n_r \ll 3n_n$, this symmetry-based reduction significantly decreases the computational cost. In particular, the computational complexity is reduced from $O(Tn_a^2)$ to $O(Tn_r^2)$, where $T$ is the total number of training time points.

\subsection{Symmetry-Reduced PINN}
\begin{figure*}
\centering
    \includegraphics[width=0.97\textwidth]{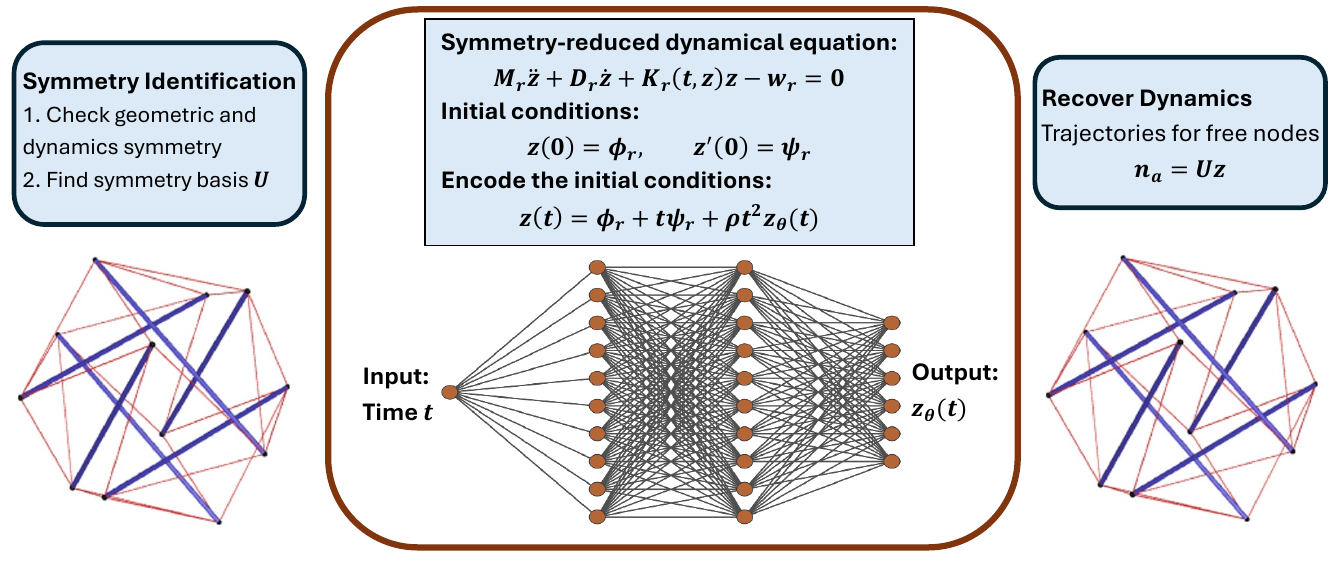}
    \caption{Pipeline of the proposed SymPINN framework. The neural network learns the dynamics in a symmetry-reduced subspace using Fourier feature embeddings, and the full free nodal trajectories are reconstructed by mapping the reduced solution back to the original space.}\label{fig:pipeline}
    \vspace{-10pt}
\end{figure*}

In this section, we introduce a symmetry-reduced physics-informed learning framework with hard constraints, referred to as SymPINN.
The overall pipeline of the proposed SymPINN framework is illustrated in Fig.~\ref{fig:pipeline}.
It consists of several steps, including symmetry detection, construction of the symmetry basis, preparation of dimension-reduced data under the symmetry transformation, training a neural network using the reduced data and the reduced governing equation, and reconstruction of the full solution in the original coordinate space.

In practice, we identify the geometric symmetry of a tensegrity structure by examining the connectivity patterns of its strings and bars. Specifically, we analyze how the nodes are connected by these structural elements and determine whether a permutation of the nodes preserves the connectivity relationships among them. If a node permutation maps each string to another string and each bar to another bar with the same connectivity pattern, then this permutation is considered a symmetry of the structure. In other words, the structural graph defined by the bar and string connections remains unchanged under such a transformation.

To implement this procedure, we first inspect the connectivity matrices that describe the bar and string elements of the tensegrity system. By comparing the connectivity patterns before and after candidate node permutations, we can identify transformations that leave both the bar and string networks invariant. These permutations indicate that the corresponding nodes belong to the same symmetry orbit and therefore share identical structural roles within the tensegrity framework. Once these symmetric node groups are determined, they can be used to construct a reduced representation of the system in which only the independent node coordinates are retained. In particular, we create a symmetry basis matrix $\bm U$ for the symmetry transform defined in \eqref{eqn:S}. This symmetry identification step plays a crucial role in enabling the symmetry-reduced modeling framework used in our approach.

Next, we describe how to incorporate symmetry into the design of a physics-informed neural network. In the standard formulation of PINNs \cite{xu2025predicting}, the training objective is typically constructed as a weighted combination of several loss components. These components usually include a physics loss that measures the residual of the governing differential equations, a data-fitting loss that enforces agreement between the network predictions and the available observations, and additional terms that impose the initial or boundary conditions of the dynamical system. Together, these loss terms guide the neural network to learn solutions that are consistent with both the observed data and the underlying physical laws. However, for second-order dynamical systems, computing higher-order derivatives of the neural network can amplify numerical errors and increase computational cost, making training more challenging. to properly balance the weights of these loss terms during training.
In the meantime, many tensegrity structures possess intrinsic geometric symmetries that can be exploited to reduce computational complexity by reducing the number of independent variables. Motivated by this observation, we integrate symmetry into a physics-informed learning framework and enforce the initial conditions through a hard constraint. Specifically, we assume the solution of the reduced system \eqref{eqn:dyn_sym} takes the form
\begin{equation}\label{eqn:soln_z}
\bm z(t)=\bm z_0 + t\,\dot{\bm z}_0 + \rho t^2 \bm z_{\theta}(t),
\end{equation}
where $\bm z_0$ and $\dot{\bm z}_0$ are defined in \eqref{eqn:dyn_sym_ic}, and the tuning parameter $\rho>0$ controls the convergence speed with a typical range $[1,100]$. The function $\bm z_{\theta}(t)$ is represented by a neural network parameterized by $\theta$, which learns the remaining temporal dynamics of the reduced coordinates. By differentiating both sides of \eqref{eqn:soln_z} with respect to $t$, we obtain
\begin{align}
\dot{\bm{z}}(t)
&=\dot{\bm{z}}_0+2\rho t\,\bm z_{\theta}(t)+\rho t^2\,\dot{\bm z}_{\theta}(t)\\
&\approx \dot{\bm{z}}_0+2\rho t\,\bm z_{\theta}(t),\label{eqn:z_t}
\end{align}
where $\dot{\bm z}_{\theta}(t)$ denotes the time derivative of the neural network output.
When $t$ is small, the term $\rho t^2\,\dot{\bm z}_{\theta}(t)$ can be neglected with minimal loss of accuracy, thereby reducing the computational cost.
This simplification is particularly beneficial because evaluating neural network derivatives via automatic differentiation (autograd) based on computational graph and chain rule is typically computationally intensive; consequently, the resulting formulation requires only a single autograd evaluation when computing the physics loss.

Given a set of training time points $\{t_i\}_{i=0}^T$, where $t_0=0$ is the initial time, we partition them into two subsets. The first subset, denoted by $T_G$, contains $N_G$ collocation time points used to evaluate the physics-based loss associated with the governing equations. The second subset, denoted by $T_D$, contains $N_D$ measurement time points used to compute the data-driven loss. In general, the sets $T_G$ and $T_D$ may overlap or be disjoint. In this work, we focus on the case where the collocation points coincide with the measurement points, i.e., $T_G = T_D$.

Let $\bm z_{\theta}(t_i)$ denote the solution of the reduced system \eqref{eqn:dyn_sym} predicted by the neural network with parameters $\theta$ at time $t_i$, and let $\bm z(t_i)$ denote the corresponding reference solution, which is either exact or measured. The training loss for the proposed SymPINN is defined as
\[
L=\lambda_G L_G+\lambda_D L_D,
\]
where $L_G$ represents the physics-based loss associated with the residual of the reduced dynamical system \eqref{eqn:dyn_sym}, and $L_D$ denotes the data-driven loss measuring the discrepancy between the predicted solution $\bm z_{\theta}$ and the reference solution $\bm z$. The weighting parameters $\lambda_G$ and $\lambda_D$ are used to balance the contributions of the two loss terms. In practice, one of these parameters can be fixed to one to reduce the tuning burden. The two loss terms are defined as
\begin{align}
L_G &= \frac{1}{N_G}\sum_{t\in T_G}
\norm{\bm{M}_r \ddot{\bm{z}}_{\theta}
+\bm{D}_r \dot{\bm{z}}_{\theta}
+\bm{K}_r\bm{z}_{\theta}
-\bm{w}_r}^2,\\
L_D &= \frac{1}{N_D}\sum_{t\in T_D}
\norm{\bm z_{\theta}(t)-\bm z(t)}^2.
\end{align}
In the definition of the physics loss, we consider the symmetry-reduced governing equation \eqref{eqn:dyn_sym} for free nodes. We adopt the same fully connected neural network architecture as in \cite{xu2025predicting}, where the neural network is implemented as a multilayer perceptron (MLP) with several hidden layers and nonlinear activation functions. The key difference in our formulation is that the initial conditions $\bm z_0$ and $\dot{\bm z}_0$ are explicitly incorporated into the network through the hard-constraint representation of the solution \eqref{eqn:soln_z} and using the approximated version of \eqref{eqn:z_t} with $\dot{\bm z}_{\theta}(t)$ omitted to enhance the computational efficiency. In this way, the neural network only learns the remaining component $\bm z_\theta(t)$, while the initial conditions are automatically satisfied exactly by construction.

To improve the expressiveness of the neural network for representing time-dependent dynamics, we incorporate Fourier feature encoding \cite{tancik2020fourier} into the SymPINN framework. Fourier features are commonly used to mitigate the spectral bias of neural networks and enhance their ability to approximate high-frequency components of the solution. Instead of directly feeding the time variable $t$ into the neural network, we first map it into a higher-dimensional feature space using sinusoidal embeddings. Specifically, we define the Fourier feature mapping to be
\begin{equation}
\gamma(t)=\begin{bmatrix}
        \sin(2\pi \omega_1 t)\\ \cos(2\pi \omega_1t)\\ \vdots\\
        \sin(2\pi \omega_K t)
        \\ \cos(2\pi \omega_K t)
    \end{bmatrix},
\end{equation}
where $\{\omega_k\}_{k=1}^K$ are the frequency parameters randomly chosen from a Gaussian distribution. The encoded feature vector $\gamma(t)$ is used as the input of the neural network that predicts the reduced coordinates
\[
\bm z_{\theta}(t)=\mathcal{N}_{\theta}(\gamma(t)),
\]
where $\mathcal{N}_{\theta}$ represents the neural network with parameters $\theta$. The full nodal coordinates are reconstructed using the symmetry basis $\bm U$ through \eqref{eqn:rec_n}. Through our extensive numerical experiments, it can be observed that using Fourier features improves the ability of the neural network to capture oscillatory behaviors and sharp temporal variations that frequently arise in tensegrity dynamics. This is particularly beneficial for the SymPINN framework, where the reduced coordinates
$\bm z(t)$
may still exhibit complex temporal patterns even after symmetry reduction.

For training, we minimize the total loss using a two-stage optimization strategy commonly adopted in physics-informed learning. In the first stage, the parameters of the neural network are optimized using the Adam optimizer for the first 1000 epochs, which provides efficient stochastic gradient updates and helps the training process quickly approach a good region of the parameter space. In the second stage, we switch to the L-BFGS optimizer to further refine the solution. The quasi-Newton updates in L-BFGS typically lead to faster convergence and improved accuracy by effectively exploiting curvature information of the loss landscape, thereby helping the optimization escape shallow local minima and achieve a more accurate approximation of the tensegrity dynamics. Our experiments have shown that L-BFGS can greatly accelerate data loss decay and improve prediction accuracy.

\section{Numerical Experiments}\label{sec:exp}

In this section, we conduct numerical experiments to evaluate the performance of the proposed SymPINN method on two datasets, namely the T-bar and the lander structures. For each dataset, we split the data into two subsets: a training set and a testing set. The training data consist of collocation and measurement points obtained through uniform random sampling, while the remaining data points are used for testing and performance evaluation. For simplicity, the collocation points are taken to be the same as the measurement points, which is appropriate since the tensegrity dynamics are smooth and the symmetry reduction reduces the effective dimensionality of the system. The sampling ratios considered in the experiments are $\{0.1, 0.2, 0.3, 0.4, 0.5\}$. To avoid the imbalance of physics loss and data loss, we divide the physics loss by the Frobenius norm of the matrix $\bm K_r$. Unless otherwise specified, the parameters are set as $\lambda_G=1$ and $\lambda_I=10$. In fact, the final performance in our experiments is not very sensitive to the choice of these two parameters. In the SymPINN architecture, we employ two hidden layers with layer size 32 for both the T-bar and the lander cases, and let $\rho=20$ in \eqref{eqn:z_t}. During the optimization procedure, the Adam optimizer is executed for a maximum of 1000 epochs, followed by 10 iterations of the L-BFGS algorithm for both datasets. For each optimizer, the iterations terminate when the relative change in the total loss falls below the tolerance $\varepsilon=10^{-5}$. Let the matrix $\bm{Y}=[\bm{n}_{t_{i_1}},\ldots,\bm{n}_{t_{i_T}}]$ contain the actual solutions evaluated at the testing time points $t_{i_1},\ldots,t_{i_T}$, and $\bm{Y}_{\theta}=[\widehat{\bm{n}}_{t_{i_1}},\ldots,\widehat{\bm{n}}_{t_{i_T}}]$ be the predicted version of $\bm{Y}$ by our method. To evaluate the prediction accuracy quantitatively, we use the following two metrics.
\begin{enumerate}[(a)]
\item Mean Squared Error (MSE) between the true solution and its estimation  at the testing time points is defined as    \begin{equation}\label{eqn:mse}
    \mathrm{MSE} = \frac1{T}\sum_{j=1}^T\norm{\bm{n}_{t_{i_j}}-\widehat{\bm{n}}_{t_{i_j}}},
    \end{equation}
    where $\norm{\cdot}$ denotes the $\ell_2$-norm of a vector, i.e., the square root of the sum of the squares of its components.
\item Relative Error (RE) between the true solution and its estimation at the testing time points is defined as
\begin{equation}
    \mathrm{RE} = \frac{\norm{\bm{Y}_{\theta}-\bm{Y}}_F}{\norm{\bm{Y}}_F},
\end{equation}
where $\norm{\cdot}_F$ denotes the Frobenius norm of a matrix, i.e., the square root of the sum of all squared entries.
\end{enumerate}
One can see that RE is a scale-invariant performance metric, independent of the magnitude of the quantities involved, and therefore enables fair comparisons across datasets with different scales. However, RE may become unstable when the entries of $\bm{Y}$ are extremely small, since the denominator in its definition approaches zero. In contrast, the MSE does not involve such normalization and therefore remains numerically stable in these cases. Throughout our experiments, the two metrics exhibit similar trends, with only minor differences.

% All numerical experiments are conducted using Python 3 on a desktop computer equipped with an Intel Core i9-9960X CPU (3.10 GHz, 16 cores), 64GB of RAM, dual NVIDIA Quadro RTX 5000 GPUs, and running Windows 10 Pro. The datasets, including time points $t_i$, ground truth solutions $\bm{n}(t_i)$, the mass matrix $\bm{M}$, the damping matrix $\bm{D}$, the time-varying stiffness matrix $\bm{K}$ and the external force $\bm{f}_{ex}$ in \eqref{eqn:dyn_vec}, are generated by the TsgFEM package \cite{ma2022tsgfem} in MATLAB R2024b. For simplicity, all nodes are free ones in the two datasets and the material type of bars and strings fixed as linear elastic. Table~\ref{tab:config} summarizes the material configurations for the datasets.
All numerical experiments are conducted in Python 3 on a desktop computer equipped with an Intel Core i9-9960X CPU (3.10 GHz, 16 cores), 64 GB RAM, and dual NVIDIA Quadro RTX 5000 GPUs running Windows 10 Pro. The datasets, including time points $t_i$, ground-truth solutions $\bm{n}(t_i)$, mass matrix $\bm{M}$, damping matrix $\bm{D}$, time-varying stiffness matrix $\bm{K}$, and external force $\bm{f}_{ex}$ in \eqref{eqn:dyn_vec}, are generated using the TsgFEM package \cite{ma2022tsgfem} in MATLAB R2024b. For simplicity, all nodes are treated as free nodes in both datasets, and the bars and strings are modeled as linear elastic materials. Table~\ref{tab:config} summarizes the material configurations.

\begin{table}
\centering
    \begin{tabular}{c|cc}\hline\hline
    Dataset & T-bar & lander \\ \hline
    bar material & wood & steel \\
    string material & rubber band & steel string \\
    %external force & step & impulse\\
    Young's modulus of bars & $8.10\times10^9$ Pa & $2.06\times 10^{11}$ Pa\\
    Young's modulus of strings & $2.00\times 10^6$ Pa& $7.60\times 10^{10}$ Pa\\
    % stress of bar & &\\
    % stress of string & & \\
    % density of bar & & \\
    % density of string & & \\
     \hline\hline    \end{tabular}
    \caption{Simulation configurations for each dataset.}\label{tab:config}
    \vspace{-10pt}
\end{table}

\subsection{Experiment 1: T-Bar}
\begin{figure}[h]   \centering\includegraphics[width=.45\textwidth]{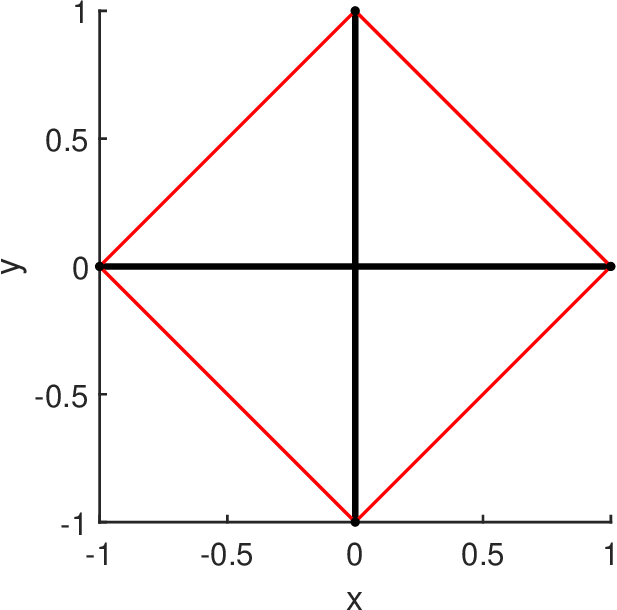}
\caption{Initial configuration of the T-bar. The black and red lines represent bars and strings, respectively. The bar length is $1$ m.} \label{fig:tbar_tsg}
\end{figure}

\begin{figure}[h]
\centering
    \includegraphics[width=.6\textwidth]{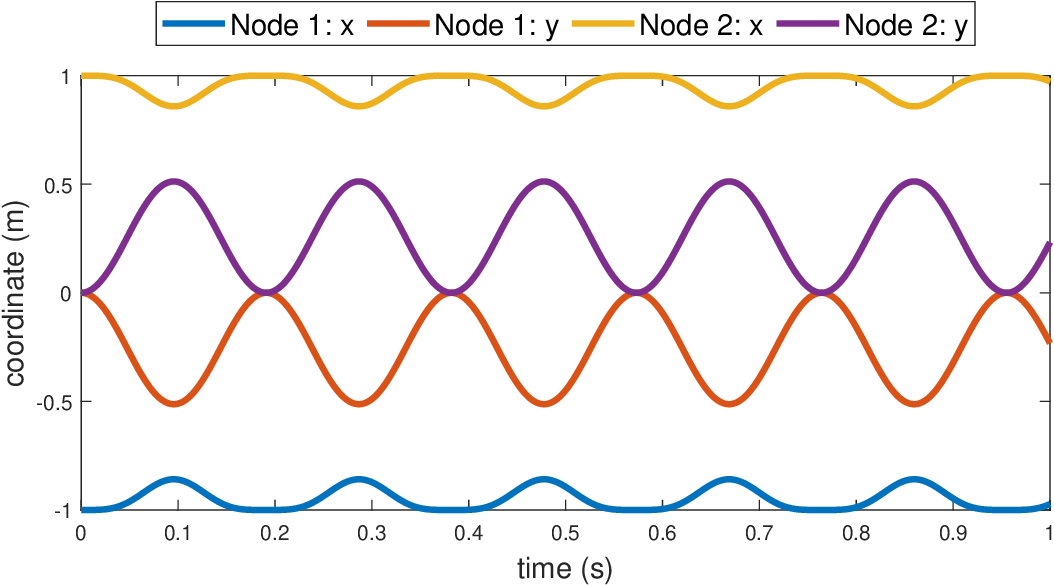}
    \caption{Trajectories of the $x$ and $y$ coordinates of the two orbit nodes in the T-bar structure.}\label{fig:tbar_nodes}
    \vspace{-10pt}
\end{figure}

In the first experiment, we consider the T-bar dataset consisting of four free nodes (with no constrained nodes). The T-bar tensegrity belongs to the $\mathcal{C}_2$ symmetry group defined in Section~\ref{sec:c_sym}, provided the structure is arranged in the standard symmetric configuration.
Using symmetry detection, we obtain the corresponding permutation and rotation matrices
\[
\bm P=\begin{bmatrix}
0&0&1&0\\
0&0&0&1\\
1&0&0&0\\
0&1&0&0
\end{bmatrix},\quad
\bm R=\begin{bmatrix}
-1&0&0\\
0&-1&0\\
0&0&1
\end{bmatrix}.
\]
This symmetry partitions the nodes into two orbits. Consequently, the dimension of the solution vector for the free nodes, originally $n_a=12$, can be reduced to $n_r=6$.
The corresponding symmetry basis matrix is denoted by $\bm U\in\mathbb{R}^{12\times 6}$. Specifically, the matrix $\bm U$ is given by
\[
\bm U=\frac1{\sqrt{2}}\begin{bmatrix}
    \bm I_{6}\\
    \bm I_2\otimes \bm R
\end{bmatrix}\in\R^{12\times 6}.
\]
The normalization constant $1/\sqrt{2}$ is added to guarantee that $\bm U^T\bm U=\bm I_{6}$. Therefore, each free nodal vector at time $t$ can be expressed by
\[
\bm n_a(t)=\frac1{\sqrt{2}}\begin{bmatrix}
    \bm z(t)\\
    (\bm I_2\otimes \bm R) \bm z(t)
\end{bmatrix},
\]
where $\bm z(t)$ is trained through the neural network. To satisfy the symmetry assumptions, we centralize the original dataset at 1001 time points evenly distributed over the time interval $[0,1]$ seconds by subtracting the nodal and velocity centroids. The initial node configuration of the T-bar is shown in Fig.~\ref{fig:tbar_tsg}. In Fig.~\ref{fig:tbar_nodes}, we plot the trajectories of the $x$ and $y$ coordinates of the two orbit nodes over the given time period, namely node 1 and node 3, in the T-bar structure. Here, the $z$ coordinate is omitted since it remains zero throughout the motion. One can see that the resulting trajectories have periodic behaviors.

\begin{figure}[h]
\centering
    \includegraphics[width=0.65\textwidth]{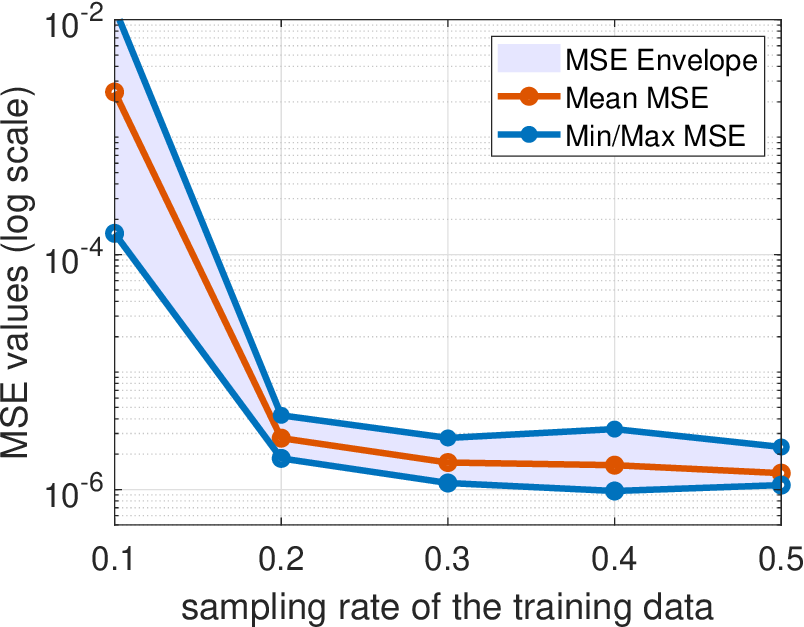}
\caption{Prediction MSE on the T-bar dataset for different training data sampling rates.}\label{fig:tbar_mse}
\end{figure}

\begin{figure}[h]
\centering
    \includegraphics[width=0.65\textwidth]{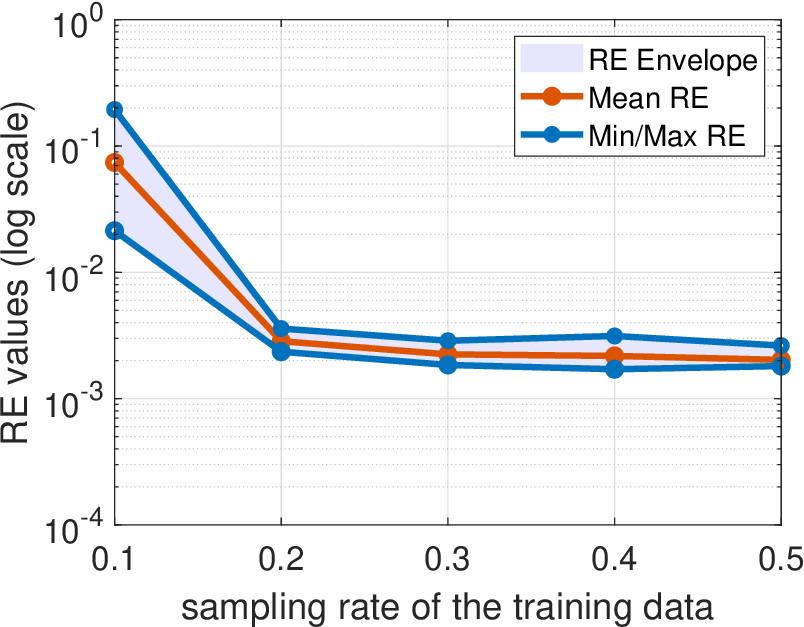}
    \caption{Prediction RE on the T-bar dataset for different training data sampling rates.}\label{fig:tbar_re}
    \vspace{-10pt}
\end{figure}

In addition, the MSE and RE values of the predicted trajectories at the testing time points under different training data sampling rates are reported in Figs.~\ref{fig:tbar_mse} and \ref{fig:tbar_re}, respectively. From these figures, it can be observed that the prediction errors decrease consistently as the amount of training data increases. In particular, there is a notable improvement in accuracy when the sampling rate increases from 10\% to 20\%, indicating that a small increase in the available training data can significantly enhance the predictive performance of the model. As the sampling rate continues to grow, the errors further decrease and gradually stabilize, which implies that the proposed model is capable of achieving reliable predictions even with a relatively limited amount of training data. The smaller variance for large sampling rates, especially in RE, indicates stability and robustness of SymPINN.

Furthermore, we compare the proposed SymPINN with the standard PINN method \cite{xu2025predicting} under identical neural network architectures and optimization settings to ensure a fair comparison. Note that the maximum number of function evaluations in each L-BFGS step is reduced from 500 to 50 to decrease runtime, while maintaining comparable accuracy. The average performance on the T-bar dataset, measured in terms of prediction accuracy and runtime, is summarized in Table~\ref{tab:tbar_perf}. Since MSE and RE have similar trends across all experiments, we report only RE in the table for a more concise presentation. In addition, the RE of PINN can be improved by increasing the network depth or width, or by allowing more function evaluations in the L-BFGS optimizer. However, this significantly increases the computational cost. For example, allowing up to 500 function evaluations in L-BFGS may reduce the RE to below $0.01$ with a small probability, but each training run would take more than 20 minutes.

Table~\ref{tab:tbar_perf} shows that SymPINN consistently outperforms PINN in terms of both prediction accuracy and computational efficiency under various training data sampling rates. In particular, SymPINN achieves significantly lower relative errors while also requiring less runtime compared to the standard PINN under the same experimental setup. This improvement can be interpreted by the symmetry-reduced representation, which effectively decreases the dimensionality of the learning problem and allows the neural network to focus on the essential degrees of freedom of the tensegrity dynamics. As the sampling rate increases, the runtime naturally increases for both methods because a larger number of training data points leads to higher computational costs during optimization. We also observe that for PINN, the Adam optimization often converges before reaching 1000 epochs, and the subsequent L-BFGS iterations provide only limited improvement in reducing the data loss. However, for SymPINN, the L-BFGS refinement plays a much more significant role, leading to further substantial decay of the loss after the Adam stage. SymPINN is able to capture the long-time periodic motion better than the vanilla PINN.
Nevertheless, SymPINN maintains a clear advantage over PINN across all the sampling rates.

\begin{table}[t]
\centering
\begin{tabular}{c|cc|cc}
\hline\hline
\multirow{2}{*}{Rate} & \multicolumn{2}{c|}{PINN} & \multicolumn{2}{c}{SymPINN}\\
\cline{2-5}
 & RE & Time (s) & RE & Time (s) \\
\hline\\[-11pt]
0.1 & 0.1850 & 272.4 & $7.40\times10^{-2}$ & 48.01 \\
0.2 & 0.1813 & 347.7 & $2.85\times10^{-3}$ & 61.66 \\
0.3 & 0.1817 & 392.2 & $2.24\times10^{-3}$ & 81.24 \\
0.4 & 0.1769 & 446.1 & $2.18\times10^{-3}$ & 110.69 \\
0.5 & 0.1768 & 480.2 & $2.03\times10^{-3}$ & 152.24 \\
\hline\hline
\end{tabular}
\caption{Prediction performance comparison between PINN and SymPINN on the T-bar dataset under various training data rates.}
\label{tab:tbar_perf}
\vspace{-10pt}
\end{table}

\subsection{Experiment 2: Lander}
\begin{figure}[th]
%\vspace{-10pt}
\centering\includegraphics[width=.55\textwidth]{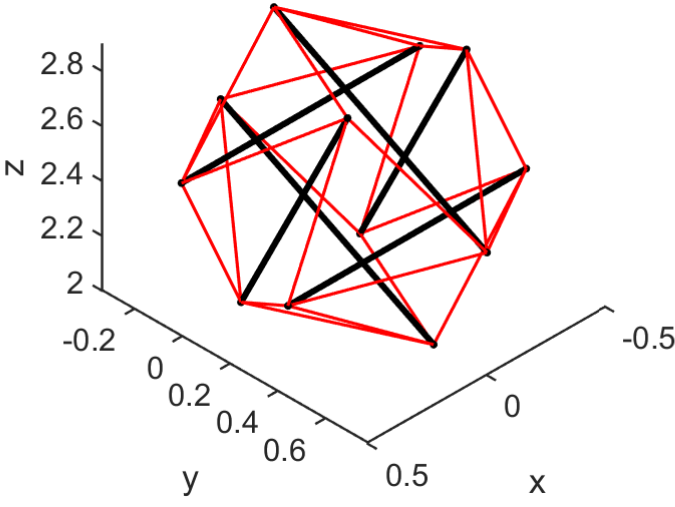}
\vspace{-6pt}
\caption{Initial configuration of the six-bar tensegrity lander. The bar length is 1 m.}\label{fig:lander_tsg}
\vspace{-10pt}
\end{figure}

\begin{figure}[h]
\centering
   \includegraphics[width=.6\textwidth]{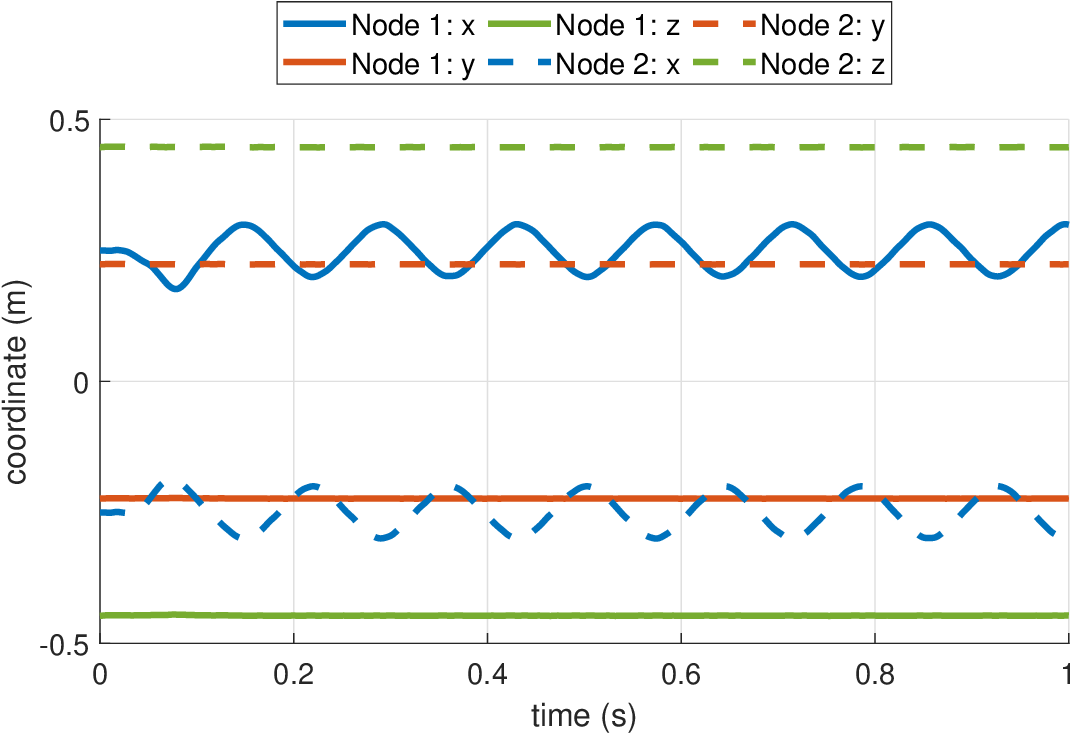}
    \caption{Trajectories of the $x$, $y$ and $z$ coordinates of the two nodes in the lander structure.}\label{fig:lander_nodes}
    \vspace{-10pt}
\end{figure}

In the second experiment, we consider the lander dataset, which consists of 12 free nodes. The lander tensegrity also belongs to the $\mathcal{C}_2$ symmetry group in Section~\ref{sec:c_sym}, implying that there are two orbits and the number of variables can be reduced from 36 to 18 in our SymPINN. The permutation matrix $\bm P$ is constructed according to the following permutation of node indices:
\[
\bm p=(3,4,1,2,6,5,8,7,12,11,10,9).
\]
The corresponding rotation matrix is given by
\[
\bm R=\begin{bmatrix}
    -1&0&0\\0&-0.6&0.8\\0&0.8&0.6
\end{bmatrix}.
\]
To obtain the symmetry basis numerically, we first construct the symmetry constraint matrix associated with the permutation–rotation symmetry. Specifically, we define the symmetry transform matrix as
$
\bm S=\bm I_6 \otimes \bm R-\bm P\otimes \bm I_3
$.
Next, we perform the Singular Value Decomposition (SVD) of $\bm S$ and let the numerical rank of $\bm S$ be determined by the number of singular values exceeding a small tolerance, e.g., $10^{-6}$. That is, we determine the numerical rank of $\bm S$ by
${r=\{\sigma_i:\sigma_i>10^{-6}\}}$.
Then the symmetry basis matrix $\bm U$ is constructed from the last $(36-r)$ right singular vectors corresponding to the singular values smaller than $10^{-6}$.
The initial velocity is applied along the $z$-axis. To ensure consistency with the structural symmetry, the external force
$\bm f_{ex}\in\mathbb{R}^{36}$ at each time point is constructed by first
reshaping the original time-varying external force
$\bm f_{0,ex}(t)\in\mathbb{R}^{36}$ into a matrix
$\bm F_{0,ex}(t)\in\mathbb{R}^{n_a\times 3}$. The force is then symmetrized
by applying the symmetry transformation
\[
\bm{F}_{ex}(t)=\frac{1}{2}\left(\bm{F}_{0,ex}(t)+\bm P\,\bm{F}_{0,ex}(t)\bm R\right).
\]
The resulting matrix $\bm F_{ex}(t)$ is subsequently reshaped back into a
vector $\bm f_{ex}(t)\in\mathbb{R}^{36}$ for use in the dynamic system.
The initial nodal configuration of the lander structure is illustrated in
Fig.~\ref{fig:lander_tsg}.
In Fig.~\ref{fig:lander_nodes}, we present the trajectories of all coordinate components for two representative nodes in the lander tensegrity structure, namely node 1 and node 12, after removing the centroid motion.

\begin{figure}[h]
\centering
    \includegraphics[width=0.65\textwidth]{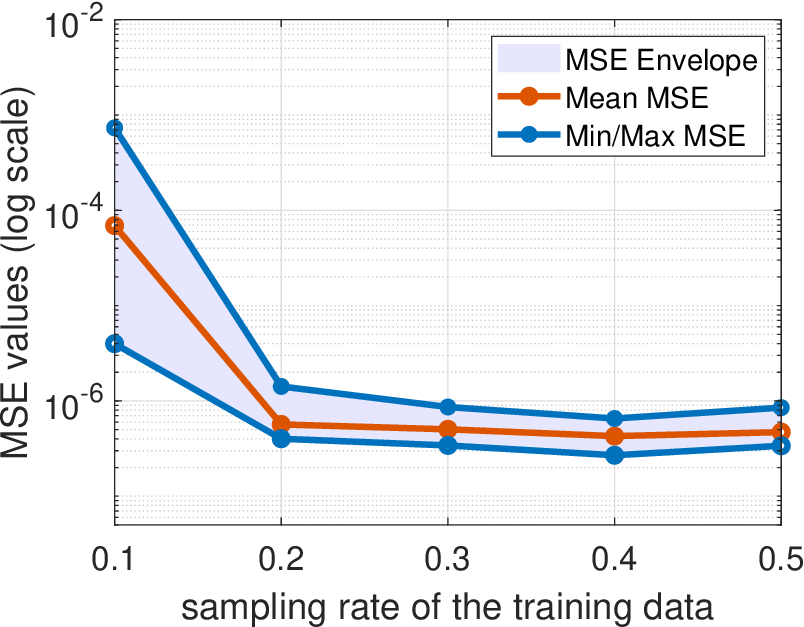}
\caption{Prediction MSE on the lander dataset for different training data sampling rates.}\label{fig:lander_mse}
\end{figure}

\begin{figure}[h]
\centering
    \includegraphics[width=0.65\textwidth]{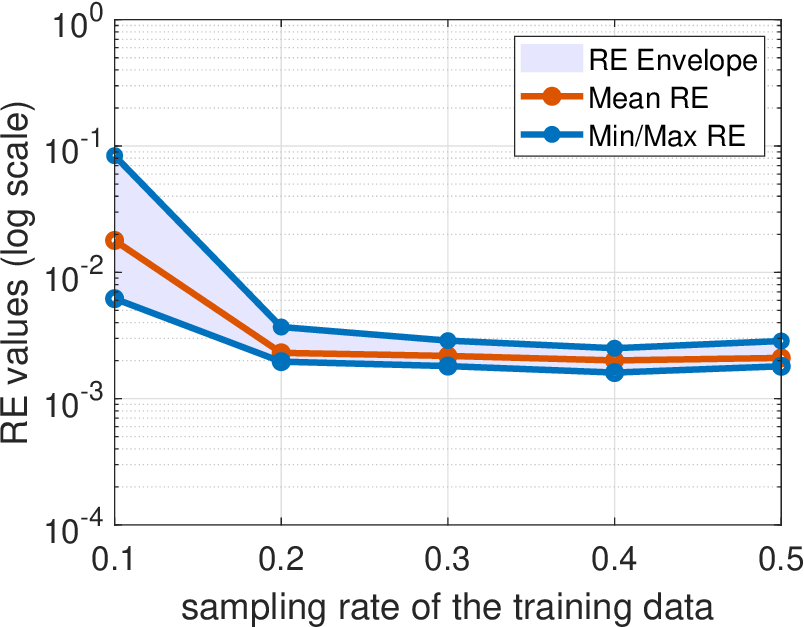}
    \caption{Prediction RE on the lander dataset for different training data sampling rates.}\label{fig:lander_re}
    \vspace{-10pt}
\end{figure}

Figures~\ref{fig:lander_mse} and \ref{fig:lander_re} report the respective MSE and RE values for the lander dataset under different training data sampling rates. Similar to the observations for the T-bar dataset, both error metrics exhibit a pronounced decrease when the sampling rate increases from $0.1$ to $0.2$. This indicates that even a modest increase in the number of training samples can significantly enhance the model's ability to capture the underlying tensegrity dynamics. As the sampling rate continues to increase beyond $0.2$, the improvements in both MSE and RE gradually diminish, and the errors become stabilized, suggesting that the model has already learned an accurate approximation of the system dynamics with a moderate amount of training data. Under the same neural network architecture and optimization settings, the lander dataset generally achieves lower prediction errors than the T-bar dataset. In particular, the MSE values for the lander dataset fall below $10^{-6}$ once the sampling rate exceeds $0.1$, indicating highly accurate reconstruction of the nodal trajectories. This behavior suggests that the lander system may exhibit smoother dynamical behavior or a more favorable symmetry structure, allowing the proposed SymPINN framework to learn the governing dynamics more efficiently even with limited training data.

Finally, we compare the performance of PINN and SymPINN on the lander dataset in terms of relative error and runtime in Table~\ref{tab:lander_perf}. Similar to the T-bar case, we use the same neural network configuration and two-phase optimization procedure but with a reduced maximum number of function re-evaluations in L-BFGS from 500 to 50 to make the runtime manageable.
SymPINN demonstrates significantly improved efficiency over the vanilla PINN in both prediction accuracy and computational cost. In particular, the L-BFGS optimization in the standard PINN becomes substantially more expensive as the number of nodes increases, due to the higher dimensionality of the solution and the associated intensive computation of physics loss. Under the same L-BFGS setting used in our SymPINN, one training run of the vanilla PINN would take several hours to complete. In the meantime, since the dimension of the solution in the reduced system is reduced by half and expensive graph-based automatic differentiation is only applied once rather than twice, SymPINN achieves faster training and improved prediction accuracy.
These results highlight the strong potential of SymPINN for modeling the dynamics of large symmetric tensegrity systems.

\begin{table}[t]
\centering
\begin{tabular}{c|cc|cc}
\hline\hline
\multirow{2}{*}{Rate} & \multicolumn{2}{c|}{PINN} & \multicolumn{2}{c}{SymPINN}\\
\cline{2-5}
 & RE & Time (s) & RE & Time (s) \\
\hline\\[-10pt]
0.1 & $6.84\times 10^{-2}$ & 2419.48 & $1.79\times10^{-2}$ & 113.13 \\
0.2 & $6.77\times 10^{-2}$ & 2634.08 & $2.30\times10^{-3}$ & 153.42 \\
0.3 & $6.88\times 10^{-2}$ & 3068.93 & $2.18\times10^{-3}$ & 226.37 \\
0.4 & $6.81\times 10^{-2}$ & 3068.56 & $2.01\times10^{-3}$ & 341.22 \\
0.5 & $6.72\times 10^{-2}$ & 2797.26 & $2.10\times10^{-3}$ & 515.33 \\
\hline\hline
\end{tabular}
\caption{Prediction performance comparison between PINN and SymPINN on the lander dataset under various training data rates.}
\label{tab:lander_perf}
\vspace{-10pt}
\end{table}
%%%%% Conclusions %%%%%%%%%%%%%%%%%%%%%%%%%%%%%%%

\section{Conclusion}\label{sec:con}

In this paper, we developed a symmetry-reduced physics-informed neural network framework, termed SymPINN, for learning the dynamics of tensegrity structures with geometric symmetry, especially cyclic and dihedral symmetries. By exploiting the inherent symmetry of the structure, the proposed method reduces the dimensionality of the governing dynamic system and enforces the symmetry constraint directly within the neural network representation.
Specifically, a symmetry basis is constructed from the nullspace of the symmetry operator corresponding to the rotation and permutation transformations of the tensegrity structure. This allows the full nodal coordinates to be expressed in terms of a reduced set of symmetry-consistent coordinates. The physics-informed neural network is then trained to approximate the evolution of these reduced coordinates, while the full configuration is reconstructed through the symmetry basis. As a result, the predicted solution automatically satisfies the geometric symmetry of the structure.

Compared with standard PINN, the proposed SymPINN framework significantly reduces the dimension of training variables and improves training efficiency. The reduced formulation also enhances numerical stability by eliminating symmetry-breaking modes that may arise during training. Moreover, initial conditions are encoded in the solution as hard constraints. Fourier features are incorporated into the neural network design, and a two-phase optimization scheme is employed to accelerate convergence and greatly improve prediction accuracy. Numerical experiments on symmetric tensegrity structures demonstrate that SymPINN can learn the system dynamics with high accuracy while requiring substantially fewer training data and neural network parameters.

The proposed framework provides a systematic way to incorporate group symmetry into physics-informed neural networks for structural dynamics. The future work includes extending this method to more general symmetry groups by integrating symmetry-aware architectures for large-scale tensegrity systems and applying this framework to other classes of mechanical systems with inherent geometric symmetry.

\bibliographystyle{unsrt}  %% .bst file following ASME conference format. Do not change.
\bibliography{ref}%% <=== change this to the name of your bib file

\end{document}